\documentclass[conference]{IEEEtran}
\IEEEoverridecommandlockouts
\usepackage{cite}
\usepackage{algorithmic}
\usepackage{graphicx}
\usepackage{textcomp}
\usepackage{xcolor}
\def\BibTeX{{\rm B\kern-.05em{\sc i\kern-.025em b}\kern-.08em
    T\kern-.1667em\lower.7ex\hbox{E}\kern-.125emX}}

\usepackage{amsmath,amssymb,amsfonts}
\usepackage{graphicx}
\usepackage{booktabs}
\usepackage{makecell}
\usepackage{subcaption}
\usepackage{fancyhdr}
\usepackage{hyperref}
\usepackage{pifont}% http://ctan.org/pkg/pifont

\usepackage[colorinlistoftodos,prependcaption,textsize=tiny]{todonotes}
\usepackage{xargs}                      % Use more than one optional parameter in a new commands
\newcommandx{\burak}[2][1=]{\todo[linecolor=green,backgroundcolor=green!25,bordercolor=green,#1]{#2}}

\title{Bit Rate Matching Algorithm Optimization in JPEG-AI Verification Model}
\author{
	\IEEEauthorblockN{
		Panqi Jia\IEEEauthorrefmark{1}\IEEEauthorrefmark{3}, A. Burakhan Koyuncu\IEEEauthorrefmark{2}\IEEEauthorrefmark{3}, Jue Mao\IEEEauthorrefmark{3}, Ze Cui\IEEEauthorrefmark{3}, Yi Ma\IEEEauthorrefmark{3}, Tiansheng Guo\IEEEauthorrefmark{3}, Timofey Solovyev\IEEEauthorrefmark{3},\\ Alexander Karabutov\IEEEauthorrefmark{3},
		Yin Zhao\IEEEauthorrefmark{3}, Jing Wang\IEEEauthorrefmark{3}, Elena Alshina\IEEEauthorrefmark{3}, André Kaup\IEEEauthorrefmark{1}}\\
	\IEEEauthorblockA{\IEEEauthorrefmark{1}Multimedia Communications and Signal Processing, Friedrich-Alexander University Erlangen-Nürnberg, Germany}
	\IEEEauthorblockA{\IEEEauthorrefmark{2}Chair of Media Technology, Technical University of Munich, Germany}
	\IEEEauthorblockA{\IEEEauthorrefmark{3}Huawei Technologies}
}

\IEEEpubid{\begin{minipage}{\textwidth}\ \\[6pt] Copyright~\copyright~2024 IEEE. Personal use of this material is permitted. However, permission to use this material for any other purposes must be obtained from the IEEE by sending an email to pubs-permissions@ieee.org\end{minipage}}

\begin{document}
\maketitle
\begin{abstract}
The research on neural network (NN) based image compression has shown superior performance compared to classical compression frameworks. Unlike the hand-engineered transforms in the classical frameworks, NN-based models learn the non-linear transforms providing more compact bit representations, and achieve faster coding speed on parallel devices over their classical counterparts. Those properties evoked the attention of both scientific and industrial communities, resulting in the standardization activity JPEG-AI.  The verification model for the standardization process of JPEG-AI is already in development and has surpassed the advanced VVC intra codec. To generate reconstructed images with the desired bits per pixel and assess the BD-rate performance of both the JPEG-AI verification model and VVC intra, bit rate matching is employed. However, the current state of the JPEG-AI verification model experiences significant slowdowns during bit rate matching, resulting in suboptimal performance due to an unsuitable model. The proposed methodology offers a gradual algorithmic optimization for matching bit rates, resulting in a fourfold acceleration and over 1\% improvement in BD-rate at the base operation point. At the high operation point, the acceleration increases up to sixfold.
\end{abstract}
% \begin{abstract}
% \end{abstract}
% \pagestyle{fancy}
% \fancyhead[ch]{}
% \fancyhead{}
%% for arxiv
% \fancypagestyle{FirstPage}{
% \lhead{\copyright  2022 IEEE. Personal use of this material is permitted. Permission from IEEE must be obtained for all other uses, in any current or future media, including reprinting/republishing this material for advertising or promotional purposes, creating new collective works, for resale or redistribution to servers or lists, or reuse of any copyrighted component of this work in other works.} 
% }
%% for arxiv
% \renewcommand{\headrulewidth}{0pt}
% \thispagestyle{FirstPage}
\begin{IEEEkeywords}
Learned Image Compression, JPEG-AI, bit rate matching, algorithm optimization
\end{IEEEkeywords}
\section{Introduction}
\noindent 
The increasing demand on online media content drives development of efficient image compression techniques, where the goal is to produce compact image representations with as high-visual quality as possible. The compression algorithms can broadly be categorized into two groups: classical and neural network (NN)-based codecs. The classical codecs, such as JPEG \cite{125072}, JPEG2000 \cite{952804}, HEVC \cite{6316136}, BPG \cite{bellard2015bpg}, and VVC intra\cite{9301847}, achieve this goal by following the pipeline of the transform coding~\cite{goyal2001theoretical}. They apply fixed transforms on image blocks, quantize the transformed vectors, and code them into a bitstream with a lossless entropy codec~\cite{rissanen1979arithmetic}.

\begin{figure}[t]
	\centerline{\includegraphics[width=0.9\linewidth]{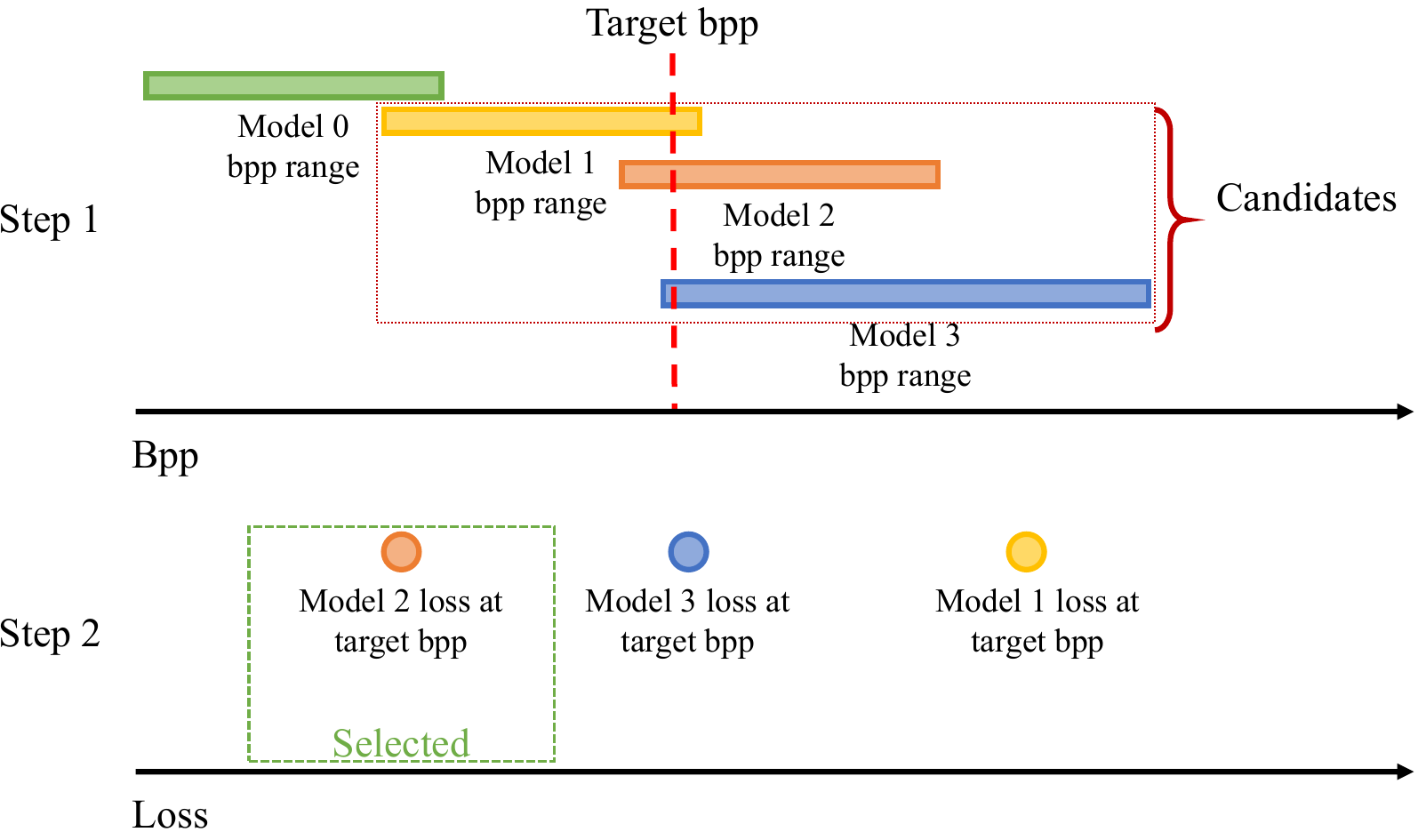}}
	\caption{Two steps of the model selection in the prior art  }
	\label{MSOld}
\end{figure}

\begin{figure*}
    \centering
    \setkeys{Gin}{width=0.3\linewidth} % <---

\subfloat[\label{ValOld}]{\includegraphics[width=0.37\linewidth]{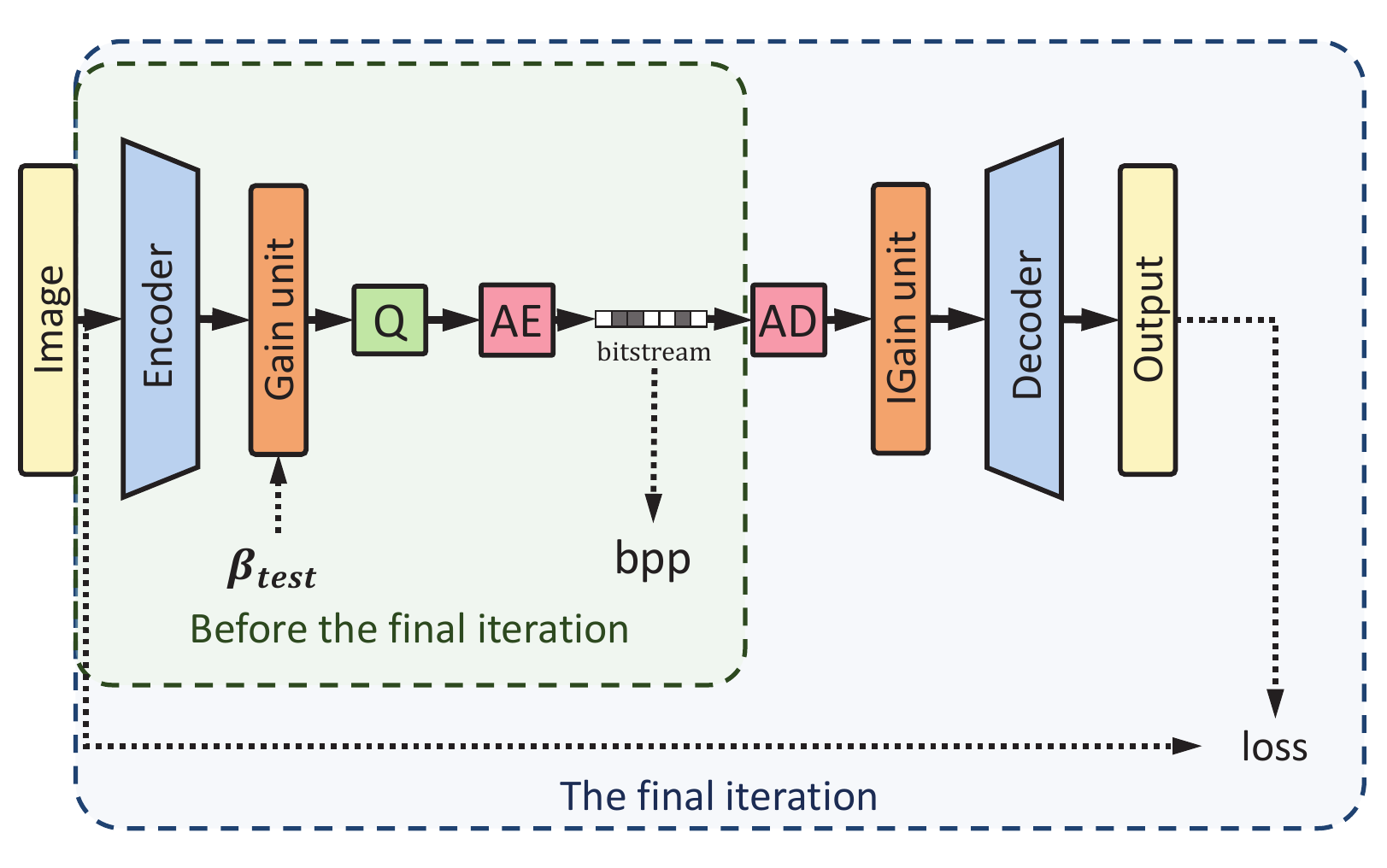} }\hfil
\subfloat[\label{NewVal}]{\includegraphics[width=0.5\linewidth]{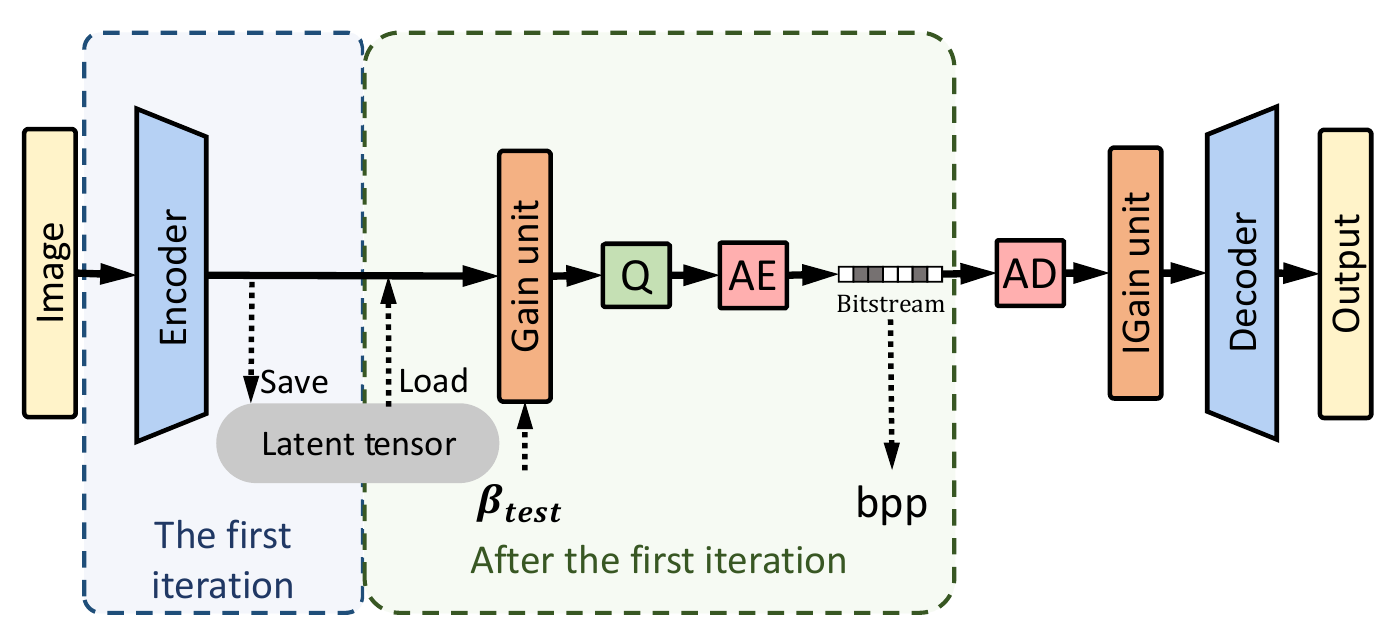} }\hfil

\caption{$\beta_{test}$ validation process (a) in prior art and (b) in our simplified method}
\label{fig:BDM_12}
\end{figure*}

In recent years, various NN-based compression algorithms have been investigated to achieve better compression performance~\cite{Ball2017EndtoendOI,Ball2018VariationalIC,Minnen2018JointAA,cheng2020learned,koyuncu2021parallelized,guo2021causal,qian2021entroformer,koyuncu2022contextformer,he2022elic,koyuncu2023efficient,liu2023learned}. Those algorithms are based on non-linear transform coding~\cite{balle2020nonlinear} using an autoencoder-based architecture with non-linear transforms, which maps the input image to a compact latent representation. The transform coefficients are learned end-to-end along with an entropy model estimating the entropy of the latent variables. Early proposals~\cite{Ball2017EndtoendOI,Ball2018VariationalIC,Minnen2018JointAA,cheng2020learned,koyuncu2021parallelized} reach a compression performance on par with classical codecs such as JPEG \cite{125072}, JPEG2000 \cite{952804}, and BPG \cite{bellard2015bpg}. More recent approaches~\cite{guo2021causal,qian2021entroformer,koyuncu2022contextformer,he2022elic,koyuncu2023efficient,liu2023learned} introduce transform layers with sophisticated activation functions~\cite{he2022elic,liu2023learned}, and entropy model techniques with attention modules~\cite{guo2021causal,qian2021entroformer,koyuncu2022contextformer,koyuncu2023efficient,liu2023learned}. Those works surpass the compression performance of VVC intra~\cite{9301847}, and even match the runtime complexity of the classical codecs~\cite{he2022elic,koyuncu2023efficient,liu2023learned}. However, they can only achieve multiple compression rates by training separate models for each rate point and cannot provide continuous variable rates with each single model during evaluation.

\IEEEpubidadjcol
To achieve continuous variable rate coding, various modifications have been proposed to the NN-based compression algorithms~\cite{choi2019variable,song2021variable,9522770,Cui2020GVAEAC,cui2021asymmetric}. For instance, Choi et al.~\cite{choi2019variable} and Song et al.~\cite{song2021variable} introduce conditional transform layers, where the transforms are adaptively modified according to the target rate. Moreover, the model of~\cite{song2021variable} can allocate more bits for the region of interest to increase coding quality on a selected location. F. Brand et al.~\cite{9522770} proposed a pyramidal NN structure with latent masking. Finetuning the mask for each layer allows for variable rate coding. Alternatively, the model of~\cite{Cui2020GVAEAC,cui2021asymmetric} proposes, a so called \textit{gain unit}, which directly adjusts the latent variables without requiring modifications to the transform layers. The gain unit proposed is a trainable matrix that contains multiple gain vectors. Each gain vector provides a channel-wise quantization map, which offers different quantization steps for each channel. Different gain vectors can provide different discrete overall quantization steps to achieve a variable rate. Furthermore, by extrapolating or interpolating the gain vectors, the codec can achieve a continuous variable rate.

The recent standardization activity, JPEG-AI~\cite{ascenso2023jpeg} has been investigating efficient NN-based codecs due the their notable compression performance. They released a verification model (VM)~4.1, where the model architecture is based on the conditional color separation (CCS) framework proposed in~\cite{jia2022learningbased}. VM~4.1 has two operation points to support different complexity requirements. The high operation point (HOP) reaches a significantly high compression performance over VVC intra~\cite{9301847}, whereas the base operation point (BOP) provides a moderate performance gain with a significantly lower model and runtime complexity. Moreover, VM~4.1 employs a gain unit similar to~\cite{Cui2020GVAEAC} in order to support continuous variable rate coding. This functionality also enables bit rate matching (BRM), where the scaling parameters for the gain unit vectors are iteratively optimized to match a given target bit per pixel (bpp). Since BRM requires an iterative algorithm, it has a significant runtime complexity. In this work, we propose novel optimization techniques for BRM, which provide up to 6.3 times lower runtime complexity and over 1\% better compression gain over the exiting BRM method.

% \section{Introduction}
\label{sec:intro}

\section{Bit Rate Matching in the JPEG-AI Verification Model}

To cover the large bpp range, JPEG-AI VM~4.1 employs five models trained for a specific Lagrange multiplier $\beta_{train}$ determining rate-distortion trade-off~\cite{balle2020nonlinear}. A higher value of $\beta_{train}$ yields a higher quality reconstructed image, requiring more bits to code. Each model uses a dedicated gain unit~\cite{Cui2020GVAEAC} to support continuous variable rate coding. To achieve the variable rate functionality for a single model, the model is provided with a test variable $\beta_{test}$ during the testing. The selection methodology for $\beta_{test}$ is explained in the following sections. Then, $\beta_{test}$ is used to compute the displacement $\delta_\beta = \frac{\beta_{test}}{\beta_{train}}$, which is the scaling parameter of gain vectors in the gain unit. This scaling parameter can assist the gain unit in extrapolating to achieve a variable overall quantization level.

In the JPEG-AI requirement, the BRM condition is considered to be satisfied, if the generated bpp differs from the target bpp by less than 10\%. In JPEG-AI VM4.1, the BRM consists of three parts: model selection, $\beta_{test}$ searching, and $\beta_{test}$ validation. In the model selection stage, BRM must locate the particular model that offers the target bpp. In the $\beta_{test}$ searching phase, the ultimate $\beta_{test}$ is determined for the selected model. In the $\beta_{test}$ validation phase, the codec computes the bpp and loss for the $\beta_{test}$. This section provides a detailed account of each BRM step.

\subsection{Model Selection}
The BRM within JPEG-AI VM4.1 selects the candidate model with the minimum loss. This model indicates the bpp range that can cover the target bpp. Different minimum and maximum $\beta_{test}$ are allocated for each model by the codec. Therefore, each model can offer a specific bpp range by utilizing the minimum and maximum $\beta_{test}$. If the target bpp is located within the bpp range of one model, then that model will be chosen as the candidate model. If the target bpp falls within the overlapping areas of multiple models' bpp ranges, then they will all be considered as candidates. Step 1 in Fig. \ref{MSOld} illustrates the chosen candidates.

Once the chosen candidate models have been identified, each model must seek out its respective $\beta_{test}$ to achieve the target bpp. This is demonstrated in step 2 of Fig. \ref{MSOld}, and the algorithm for the search for the $\beta_{test}$ is presented in subsection \ref{BetaSearch}.  Once the relevant $\beta_{test}$ is found for each candidate model, the corresponding loss of that $\beta_{test}$ is also determined. The loss is determined by comparing the reconstructed image with the original image and is a combination of the Mean Squared Error (MSE)~\cite{Sgaard2016ApplicabilityOE} and the Multi-Scale Structural SIMilarity (MS-SSIM)~\cite{1292216} loss with varying weights.
Ultimately, the model with the lowest loss and its corresponding $\beta_{test}$ will be chosen.

\subsection{$\beta_{test}$ Searching for Single Model}
\label{BetaSearch}
In the model selection, each proposed model must identify the final $\beta_{test}$ required to attain the target bpp. However, the model lacks knowledge of the ultimate destination of $\beta_{test}$. It only has access to the minimum and maximum $\beta_{test}$ values, which define the minimum and maximum bpp. Thus, the binary search is employed in this scenario. At each search iteration, the new $\beta_{test}$ is updated as $\beta_{test} = \sqrt{ \beta_{test\_max} \times \beta_{test\_min}}$. After each iteration, the new minimum and maximum $\beta_{test}$ are updated by examining the bpp range that can cover the target bpp. The revised minimum $\beta_{test}$ can generate the closest bpp to the target bpp in the bpp range that is below the target bpp. The updated maximum $\beta_{test}$ can generate the closest bpp to the target bpp in the bpp range that is above the target bpp. The $\beta_{test}$ is continually updated until it has reached the target bpp. 

\subsection{$\beta_{test}$ Validation}

Fig. \ref{ValOld} shows the validation process for $\beta_{test}$. During the $\beta_{test}$ search, several $\beta_{test}$s require validation in candidate models. Two scenarios exist for $\beta_{test}$ validation: when the $\beta_{test}$ is unable to produce the desired bpp and when the $\beta_{test}$ can supply bpp that approximates the target bpp. For both cases, the encoder must run in its entirety to obtain the latent tensor. Once the $\beta_{test}$ has been applied to the latent tensor, the modified tensor goes to the entropy model to generate the bpp. If the bpp is not sufficiently close to the target bpp, the validation will cease, and the next $\beta_{test}$ will be updated. Conversely, if the bpp is within range of the target bpp, the decoder will run to reconstruct the image for loss calculation. This loss calculation is used for selecting the model from the candidates.

\begin{table*}[t]
        \newcommand\ph{\phantom{0}}
        \newcommand\phm{\phantom{-0}}

	\centering
	\caption{Average BD rate and run time of the original (org.) and the proposed (prop.) BRM methods applied on BOP and HOP models of JPEG-AI VM4.1.}
	\label{resultTab}
	\footnotesize
	\begin{tabular}{l|c|c|c|c|c|c|c|c|c|c|c}
		\toprule
		\textbf{Model}&\textbf{AVG}&\textbf{RunTime}&\textbf{BRMTime}&\textbf{ BitDiff}&\textbf{MSSSIM}&\textbf{VIF}&\textbf{FSIM}&\textbf{NLPD}&\textbf{IWSSIM}&\textbf{VMAF}&\textbf{PSNRHVS}\\
		\midrule
		\makecell[tl]{ VTM 11.1\\ BOP (tools off) \\ $\;+$ BRM (org.) \\ $\;+$ BRM (prop.)  \\ HOP (tools off) \\ $\;+$ BRM (org.)  \\ $\;+$ BRM (prop.)  }
		
		&\makecell[t]{\phm0.0\% \\ -11.0\% \\\ph-8.1\% \\ \ph-9.2\% \\ -25.7\% \\ -22.1\% \\ -22.1\%}
		
		&\makecell[t]{-- \\ \ph29 min \\ \ph49 min \\ \ph34 min \\ \ph48 min \\ 282 min \\ \ph85 min }
		
		&\makecell[t]{-- \\ -- \\ \ph20 min \\ \ph\ph5 min \\ -- \\ 234 min \\ \ph37 min }
		
		&\makecell[t]{\ph10\% \\ 315\% \\\ph\ph6\% \\ \ph\ph4\% \\ 366\% \\ \ph\ph9\% \\ \ph\ph4\%}
		
		&\makecell[t]{\phm0.0\% \\ -29.1\% \\-26.3\% \\ -26.9\% \\ -39.2\% \\ -35.6\% \\ -35.4\%}
		
		&\makecell[t]{\phm0.0\% \\ \ph-1.7\% \\\ph-1.2\% \\ \ph-2.1\% \\ -16.8\% \\ -15.0\% \\ -15.3\%}
		
		&\makecell[t]{\phm0.0\% \\ -13.4\% \\\ph-7.5\% \\ -10.7\% \\ -27.5\% \\ -21.9\% \\ -21.9\%}
		
		&\makecell[t]{\phm0.0\% \\ -10.3\% \\\ph-6.6\% \\ \ph-8.2\% \\ -24.6\% \\ -21.1\% \\ -21.1\%}
		
		&\makecell[t]{\phm0.0\% \\ -25.1\% \\-21.7\% \\ -22.8\% \\ -36.4\% \\ -32.6\% \\ -32.4\%}
		
		&\makecell[t]{\phm0.0\% \\ \ph-1.2\% \\ \phm0.7\% \\ \phm0.6\% \\ -23.2\% \\ -19.9\% \\ -19.7\%}
		
		&\makecell[t]{\phm0.0\% \\ \phm3.4\% \\ \phm6.2\% \\ \phm5.8\% \\ -12.2\% \\ \ph-8.5\% \\ \ph-8.6\%}
		
		\\
		\bottomrule
	\end{tabular}
\end{table*}

%% for arxiv
% \fancyhead[CH]{To be presented at the Picture Coding Symposium (PCS), 7-9 December 2022, San José, California, USA}
%% for arxiv
\section{Proposed Algorithm}

In this section, we present our novel BRM optimization algorithm. Our algorithm optimises all three parts of the BRM and is introduced in the following subsections, where optimization of each part will be explained.

\subsection{Relative Bit Distance Based Model Selection}

In the field of model selection, various models are considered as candidates, and the ultimate model chosen is the one with the least loss. Nevertheless, this approach presents two drawbacks. Firstly, if there are numerous candidates, each candidate requires a $\beta_{test}$ searching process for the target bpp, which is very time-consuming. Secondly, each candidate needs to calculate the loss in the end, prolonging the runtime of the decoder and metric calculation.  

\begin{figure}[t]
	\centerline{\includegraphics[width=0.9\linewidth]{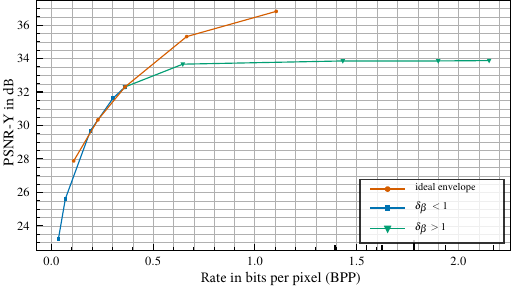}}
	\caption{Variable rate curve of a single model with different $\delta_\beta$}
	\label{beta_curve}
\end{figure}

The combination of five models with $\delta_\beta = 1$ presents the optimal variable rate envelope. If $\delta_\beta$ deviates significantly from 1, the resulting rate-distortion curve diverges from the ideal envelope. Fig. \ref{beta_curve} depicts the ideal envelope for all five models (represented by the orange curve), and the rate-distortion curve of a single model (depicted by the blue and green curve), which is trained for $\beta_{train} = 0.015$ and tested for different $\delta_\beta$ values. Notably, $\delta_\beta >1$ causes a bigger discrepancy from the ideal envelope than using $\delta_\beta < 1$. 

\begin{figure}[t]
    \centering
    \includegraphics[width = 0.72\linewidth]{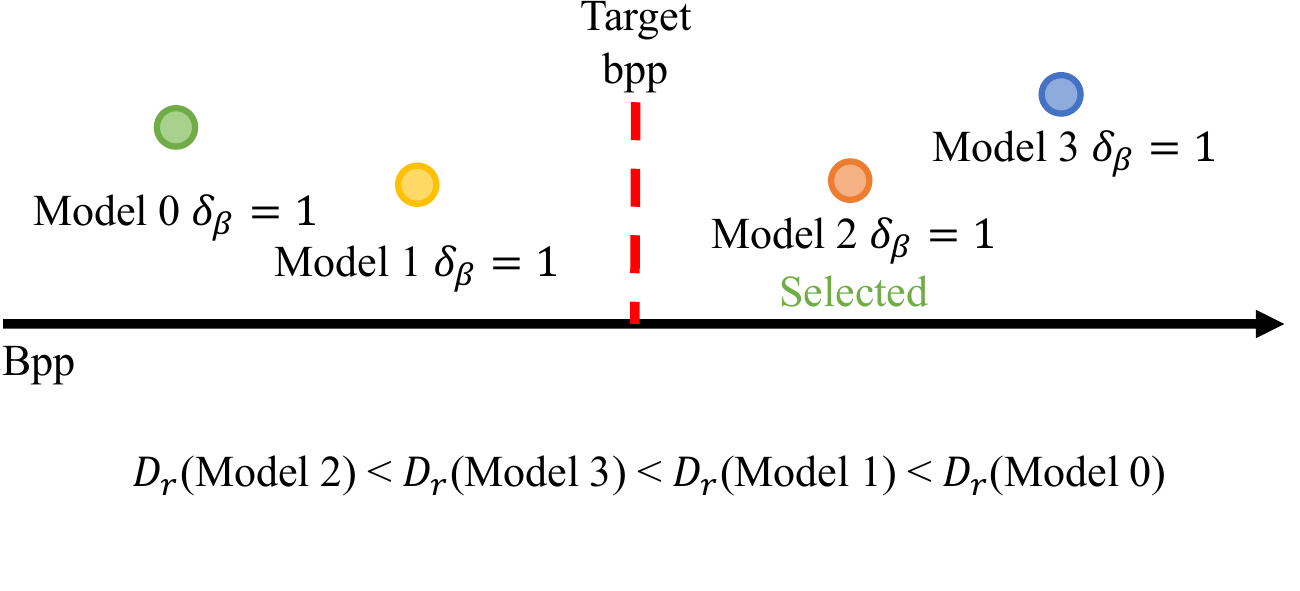}
    \caption{Directly select model by comparing the $D_r$ }
    \label{reletiveDistance}
\end{figure}

Based on these observations, the model that has the closest default bpp to the target bpp should be used to achieve the desired bpp. If the default bpp of the chosen model is higher than the target bpp, it can still provide a smaller gap from the ideal envelope. We suggest using the relative bit distance $D_r = {\rm abs}(bpp_{default} - bpp_{target})/bpp_{default}$ as the evaluation criterion. The model with the minimum $D_r$ to the target bpp will be selected directly.
For instance, as illustrated in Fig. \ref{reletiveDistance}, the target bpp lies midway between the $bpp_{default}$s of Model 1 and Model 2. The absolute difference in $bpp_{default}$ between the two models and the target bpp is identical. However, due to the higher $bpp_{default}$ of Model 2, its $D_r$ is smaller. This model selection method generates a larger lower boundary than the upper boundary for the bpp range of each model. This non-symmetric boundary can result in variable rate performance that closely approximates the ideal envelope.  

With the proposed model selection optimization, all models need to be calculated only once for the default bpp, and beta searching occurs only for the selected model. Additionally, the loss calculation is not necessary for this optimization. By simplifying the model selection strategy, significant runtime is saved.

\subsection{$\beta_{test}$ Searching Based on Linear Function}
When searching for the appropriate $\beta_{test}$ to achieve target bpp in prior art, the binary search method is utilized due to the ambiguous connection between $\beta_{test}$ and bpp. Nevertheless, our statistical analysis revealed a proper approach to model the relationship between $\beta_{test}$ and bpp.

\begin{figure}[t]
    \centering
    \includegraphics[width = 0.9\linewidth]{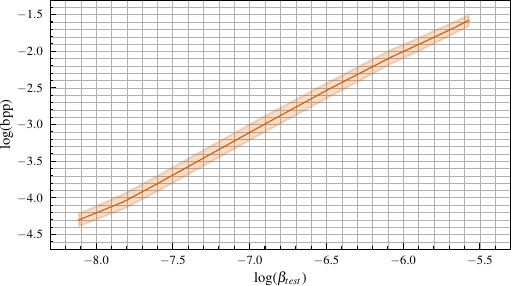}
    \caption{Relationship between ${\rm log}(bpp)$ and ${\rm log}(\beta_{test})$}
    \label{LinearRelation}
\end{figure}

In Fig. \ref{LinearRelation}, the graph displays the natural logarithm of $\beta_{test}$ and $bpp$. The logarithmic relationship between ${\rm log}(\beta_{test})$ and ${\rm log}(bpp)$ is approximately linear, allowing us to hypothesize that  ${\rm log}(bpp) = A \cdot {\rm log}(\beta_{test}) + B $. In the initial iteration of $\beta_{test}$ search, we calculate two parameters, A and B, by using the minimum and maximum $\beta_{test}$ values along with their corresponding bpp. Then, we determine $\beta_{test\_1}$ by setting the linear function of ${\rm log}(bpp) = {\rm log}(bpp_{target})$. We subsequently validate $\beta_{test\_1}$ in order to obtain $bpp_{test\_1}$. If $bpp_{test\_1}$ is close enough to the $bpp_{target}$, the search process will end. Otherwise, the linear function incorporating $\beta_{test\_min}$, $\beta_{test\_1}$ and their associated bpp will be created anew. We repeat the preceding process until reaching a value close enough to the target bpp.

Using a linear function to fit the connection between $\beta_{test}$ and bpp can decrease the number of iterations required to search for $\beta_{test}$. This approach provides a clear direction for locating the $\beta_{test}$.

\subsection{$\beta_{test}$ Validation simplification}

In the prior art, as illustrated in Fig. \ref{ValOld}, the encoder had to operate for each iteration while searching for $\beta_{test}$, and the decoder had to run upon discovering a valid $\beta_{test}$ to produce the loss. 

Noting that $\beta_{test}$ only affects the gain vector, which represents the channel-wise quantization map of the latent tensor, it follows that the latent tensor is unchanged before the gain vector is applied in each iteration. To prevent the repeated generation of the unmodified latent tensor, we suggest storing the unmodified latent tensor during the first iteration and retrieving it during ensuing iterations. This approach saves computational time for the encoder. Furthermore, our proposed approach eliminates the need to compare all possible candidates during model selection, thereby preventing the need for loss calculation. By avoiding the calculation of the loss function, we can significantly reduce the runtime required for the decoder and metric calculations. The simplified $\beta_{test}$ validation is illustrated in Fig. \ref{NewVal}.

\section{Experimental Results}

To assess the efficacy of our methodology, we compare it with JPEG-AI VM4.1. To accurately measure the BRM runtime, we initially carry out the tool-off configuration without BRM and record the runtime $T_{A}$. Next, for prior art, we enable BRM on top of the tool-off configuration and record the runtime $T_{B}$. The difference $T_{B} - T_{A}$ denotes the BRM runtime of the prior art. Our proposed method was implemented on a tool-off configuration and the resulting run time $T_{C}-T_{A}$ was recorded. To ensure consistent test conditions, we used a single NVIDIA Titan RTX graphics processing unit (GPU) with 24GB memory for all tests. Five target bpp points were evaluated: 0.06, 0.12, 0.25, 0.5, and 0.75 bpp. The models used for BRM were trained with beta values of 0.002, 0.007, 0.015, and 0.05. The $\delta_\beta$ ranges for each model were [0.1, 2.0], [0.3, 1.4], [0.4, 2], and [0.6, 6.0].
To test tool-off configuration without BRM, $\delta_\beta$ equal to 1 are used for each individual model. Furthermore, for the high rate range (0.75 bpp), we apply a $\delta_\beta$ equal to 2 for the last model. The results of the BRM are presented in Table \ref{resultTab}. The VVC Test Model (VTM 11.1) is used as the performance anchor in the table for VVC intra. Additionally, FFMPEG~\cite{tomar2006converting} will be used to convert the PNG (RGB) to YUV following the BT.709 primaries to generate the YUV format for metric calculation. 

All experiments were conducted on the JPEG-AI Common and Training Test Conditions (CTTC)\cite{wg1n100600} test dataset, comprising 50 images with natural content. To evaluate the metrics, we computed the average Bjøntegaard Delta rate (BD-Rate)~\cite{bjontegaard2001calculation} performance across 7 metrics, including MS-SSIM~\cite{1292216}, Visual Information Fidelity (VIF)~\cite{citVif}, feature similarity (FSIM)~\cite{5705575}, Normalized Laplacian Pyramid (NLPD)~\cite{nlpdcite}, Information Content Weighted Structural Similarity Measure (IW-SSIM)~\cite{5635337}, Video Multimethod Assessment Fusion (VMAF)~\cite{VMAFcite}, and PSNR-HVS-M~\cite{psnrHVS}. Furthermore, we recorded the bit difference for determining the accuracy of BRM.   

For the base operation point (BOP) of relatively lower complexity, our novel BRM exhibits a 5-minute runtime and an average BD-rate gain of -9.2\% compared to VVC intra. In contrast, the BRM of the prior art has a runtime of 20 minutes and a gain of -8.1\% over VVC. Therefore, our novel BRM achieves a 4 times increase in speed and a -1.1\% improvement in BD-Rate when compared to the prior art in JPEG-AI VM4.1 over VVC intra.  
Similarly, for the high operation point (HOP) with greater complexity, our novel BRM has a run time of 37 minutes and achieves a BD-rate gain of -22.1\% over VVC intra. In comparison, the BRM in the prior art has a run time of 234 minutes and achieves a BD-rate gain of -22.1\% over VVC intra. In summary, our novel BRM leads to a 6.3-fold enhancement in speed and attains similar average BD-rate proficiency to the prior art in JPEG-AI VM4.1 over VVC intra.

\section{Conclusion}
In this paper, we propose a novel BRM approach on top of JPEG-AI VM4.1, comprising relative bit distance-based model selection, linear function-based $\beta_{test}$ searching, and non-decoder required $\beta_{test}$ validation. 

Compared to the BRM in JPEG-AI VM~4.1 over VVC intra, the proposed BRM for the base operation point achieves a -1.1\% improvement in BD-rate and a 4-fold increase in speed. For the high operation point, the proposed BRM has an average BD-rate performance that is equivalent and provides a 6.3 times increase in speed.

\bibliographystyle{IEEEtran}
% \bibliography{strings,refs}
\bibliography{References}
% \printbibliography
\end{document}